\documentclass[sigconf]{acmart}
\usepackage{microtype}
\usepackage{graphicx}
\usepackage{booktabs}
\usepackage{multicol}
\usepackage{multirow}
\usepackage{float}
\usepackage{url}
\usepackage{makecell}
\usepackage[ruled,vlined]{algorithm2e}
\usepackage{algpseudocode}
\usepackage{subfigure}
\usepackage{mathtools}

\newcommand\retrievalname{Heterogeneous Information Retrieval}

\newcommand\generationnames{Prompt-based Generation\ }
\newcommand\modelname{LittleMu\xspace}
\newcommand\modelnames{LittleMu\ }

\algrenewcommand\algorithmicrequire{\textbf{Input:}}

\AtBeginDocument{%
  \providecommand\BibTeX{{%
    \normalfont B\kern-0.5em{\scshape i\kern-0.25em b}\kern-0.8em\TeX}}}

\setcopyright{acmcopyright}
\copyrightyear{2018}
\acmYear{2018}
\acmDOI{XXXXXXX.XXXXXXX}

\acmConference[Conference acronym 'XX]{Make sure to enter the correct
  conference title from your rights confirmation emai}{June 03--05,
  2018}{Woodstock, NY}
\acmPrice{15.00}
\acmISBN{978-1-4503-XXXX-X/18/06}

\copyrightyear{2023}
\acmYear{2023}
\setcopyright{rightsretained}
\acmConference[CIKM '23]{Proceedings of the 32nd ACM International
Conference on Information and Knowledge Management}{October 21--25,
2023}{Birmingham, United Kingdom}
\acmBooktitle{Proceedings of the 32nd ACM International Conference on
Information and Knowledge Management (CIKM '23), October 21--25, 2023,
Birmingham, United Kingdom}\acmDOI{10.1145/3583780.3615484}
\acmISBN{979-8-4007-0124-5/23/10}
\begin{document}

\title[LittleMu: Deploying an Online VTA via Heterogeneous Sources Integration and Chain of Teach Prompts]{\modelname: Deploying an Online Virtual Teaching Assistant via Heterogeneous Sources Integration and Chain of Teach Prompts}

\author{Shangqing Tu}
\authornote{Both authors contributed equally to this research.}
\email{tsq22@mails.tsinghua.edu.cn}
\author{Zheyuan Zhang}
\authornotemark[1]
\email{zheyuan-22@mails.tsinghua.edu.cn}
\author{Jifan Yu}
\email{yujf21@mails.tsinghua.edu.cn}
\affiliation{%
  \institution{Tsinghua Univerisity}
  \city{Beijing}
  \country{China}
}

\author{Chunyang Li}
\email{lichunya20@mails.tsinghua.edu.cn}
\author{Siyu Zhang}
\email{siyu-zha20@mails.tsinghua.edu.cn}
\author{Zijun Yao}
\email{yaozj20@mails.tsinghua.edu.cn}
\affiliation{%
  \institution{Tsinghua Univerisity}
  \city{Beijing}
  \country{China}
}

\author{Lei Hou}
\email{houlei@tsinghua.edu.cn}
\author{Juanzi Li}
\authornote{Corresponding author.}
\email{lijuanzi@tsinghua.edu.cn}
\affiliation{%
  \institution{Tsinghua Univerisity}
  \city{Beijing}
  \country{China}
}

\renewcommand{\shortauthors}{Trovato and Tobin, et al.}

\begin{abstract}

 Teaching assistants have played essential roles in the long history of education. However, few MOOC platforms are providing human or virtual teaching assistants to support learning for massive online students due to the complexity of real-world online education scenarios and the lack of training data. In this paper, we present a virtual MOOC teaching assistant, LittleMu with minimum labeled training data, to provide question answering and chit-chat services.
  Consisting of two interactive modules of heterogeneous retrieval and language model prompting, LittleMu first integrates structural, semi- and unstructured knowledge sources to support accurate answers for a wide range of questions. Then, we design delicate demonstrations named ``Chain of Teach'' prompts to exploit the large-scale pre-trained model to handle complex uncollected questions. Except for question answering, we develop other educational services such as knowledge-grounded chit-chat. 
We test the system's performance via both offline evaluation and online deployment. Since May 2020, our LittleMu system has served over 80,000 users with over 300,000 queries from over 500 courses on XuetangX MOOC platform, which continuously contributes to a more convenient and fair education. Our code, services, and dataset will be available at  \url{https://github.com/THU-KEG/VTA}.
\end{abstract}




\maketitle

\section{Introduction}

Teaching assistants (TAs), the senior students who assist teachers with diverse instructional responsibilities, have played essential roles in the long history of education~\cite{farrell2010impact}. In the era of Massive Open Online Courses (MOOCs), although the prosperity of online education has provided explosive amounts of learning resources for worldwide learners, it is quite difficult to retain sufficient manpower to offer detailed question and answering~\cite{goel2018jill}, learning inspiration~\cite{hone2016exploring}, and interactive instruction~\cite{hollands2014moocs} as in traditional classrooms. With the rapid development of relevant techniques (such as the large-scale pre-trained models~\cite{devlin2018bert,brown2020language}), pioneer researchers have realized that it is promising to build intelligent virtual teaching assistants (VTAs), which can provide continuous, concrete companionship and mentoring for supporting individual students~\cite{feng2019understanding}.

Despite a few previous efforts that simplify VTAs to domain-specific question answering systems~\cite{goel2018jill,hsu2022xiao} or chatbots~\cite{han2022making}, constructing an applicable VTA is still an intractable and tricky task due to the complexity of real-world online education scenarios, which can be summarized as several main technical challenges:



$\bullet$\textbf{Complexity of Integrating Heterogeneous Knowledge}. Distinct from conventional domain-specific question answering~\cite{raamadhurai2019curio,boateng2022kwame} that only requires limited types of knowledge, student queries from real MOOCs may vary across heterogeneous sources, such as platform usage (\emph{e.g.} \emph{How to participate in a previous course?}), course concepts (\emph{e.g.} \emph{What is Graph Neural Network?}) or even information seeking (\emph{e.g.} \emph{Who is the most experienced teacher in this domain?}), which poses strict requirements to the retrieval and curation of a quite wider range of structural, semi- or unstructured knowledge.

\begin{figure}[htbp] 
\centering 
\includegraphics[width=0.9\linewidth]{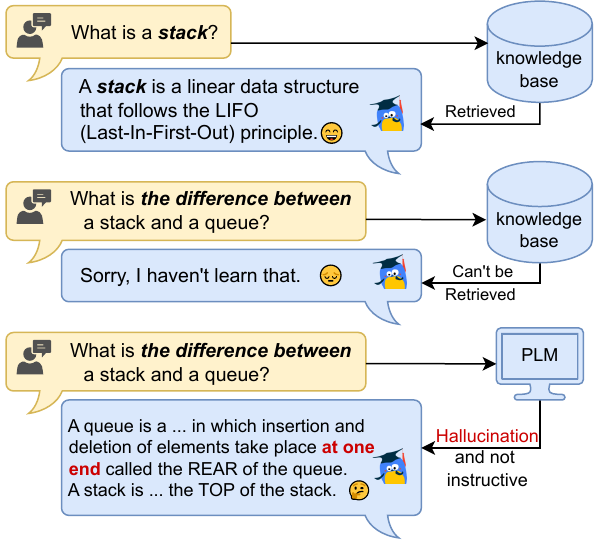} 
\caption{Complex questions can't be directly solved by simple retrieving, while generative PLMs are risky to have hallucinations and lack instructiveness.}
\label{fig:motivation_example} 
\end{figure}

$\bullet$\textbf{Difficulties in Answering Complex Questions}. Beyond the simple queries that can be directly solved via direct retrieval, a larger amount of cognitive questions that can benefit learning, are complex and rely on a series of reasoning and analyzing~\cite{wei2022chain,yu2021mooccubex}, such as \textit{Why}, \textit{How} and \textit{Comparison}, as illustrated in Figure \ref{fig:motivation_example}. Such questions, however, are hard to be fully collected in advance, thereby seriously challenging the abilities of knowledge reasoning and answer generation of an inspiring VTA.

$\bullet$\textbf{Low Transfer Ability among Courses}. As new courses are constantly emerging on the MOOC platforms every day~\cite{yu2021mooccubex}, it is costly to conduct heavy training for building VTAs in every single course~\cite{benedetto2019rexy}. Although there have been discussions about
balancing the effectiveness and transfer ability of the models~\cite{goel2022agent}, deploying a VTA that can be conveniently adopted to courses in a wide variety of subjects is a formidable topic to be explored.


To face the aforementioned challenges, we propose \modelname, an implementation of an online virtual teaching assistant which is currently providing services on over 500 courses in XuetangX\footnote{\href{https://xuetangx.com}{https://xuetangx.com}}, one of the largest MOOC platforms in China. Consisting of two interactive modules of heterogeneous retrieval and language model prompting, LittleMu preserves several major features including (1) \emph{High-coverage Q\&A}: LittleMu integrates knowledge sources such as concept-centered MOOCCubeX~\cite{yu2021mooccubex}, web search engine, platform FAQ, thereby supporting accurate and informative Q\&A for a wide range of questions;
(2) \emph{Instructional Complex Reasoning}: we design delicate demonstrations named ``Chain of Teach'' prompts to exploit the emergent reasoning ability of the large-scale pre-trained model, which enables LittleMu to handle complex uncollected questions; (3) \emph{Easy-to-adapt Transferability}: Empowered by a meta concept graph and tuning-free prompting of large models, LittleMu does not require further training stage to be applied to new courses.

We both test the performance via offline evaluation and online deployment. Except for the results from manual annotations and automatic metrics that prove the coherence, informativeness, and helpfulness of LittleMu, the increase and feedback of blind MOOC users also verify the applicability and effectiveness of our system.

\textbf{Contributions and Predicted Impact}. (1) For the researchers of knowledge-grounded dialogue and question answering, we present an example implementation that combines the retrieval-based, generative Q\&A and dialogue methods for building a practical system; (2) For the contributors of intelligent education and online learning, we conduct a series of investigations, insights, and explorations on real-world MOOCs about how to satisfy the students' interactive needs with advanced AI techniques.

LittleMu also provides a positive impact on online students: (3) Since May 2020, our proposed LittleMu has served over 80,000 users with over 300,000 queries from over 500 courses, which continuously contributes to a more convenient and fair education. We hope our work can call for more efforts in building advanced, next-generation education platforms that can benefit more learners.

\section{Preliminaries}

\subsection{Data Analysis}
\label{sec:question_analysis}

In this section, we analyze the chat history from XiaoMu, the original VTA system deployed on XuetangX since May 2020. From these queries in real MOOC scenarios, we aim to reveal their components and understand the real demands of students for a VTA.

\textbf{Data Background.}
The analysis is performed on the original queries from XuetangX, one of the largest MOOC platforms in China, equipped with more than $5,000$ courses from top universities in the world, and has attracted more than $100,000,000$ online users. 
Up to this paper, there have been more than $31,000$ dialogue sessions, $312,000$ queries from more than $590$ courses. In order to evenly collect information at different times, we applied systematic sampling to form the final data set and sampled $4,002$ questions in terms of sessions for assurance of completeness, which means all the contexts were preserved. 
We find that these questions mainly fall into three categories:
\textbf{Knowledge Questions} query for certain knowledge, \textbf{Chit-Chat} looks for small talk, and \textbf{Others} such as platform FAQs.
We hire annotators, who are teachers or MOOC frequent users, to label the questions into these categories.
The analysis results are shown in Table~\ref{tab:question_component}.

\textbf{Questions Categories.}
To dig out deeper patterns, queries are further split into more detailed sub-types. For those questions asked to obtain certain knowledge, we divide them into \textit{simple questions} and \textit{complex questions}, based on the difficulty to answer them. As shown in Table \ref{tab:question_component}, a query will be labeled as complex if it involves higher-order thinking, including comparison, method, reason, instance, recommendation and others. For those questions looking for chit-chat, we divide them into \textit{emotional} and \textit{general chit-chat}, according to whether students are asking for emotional comfort. Table \ref{tab:question_component} shows that 60\% of questions are Knowledge Questions, 12\% of which are complex, while 18\% are Chit-Chat, and 16\% of which involve affective interaction. Our observations are as follows.


\begin{table}[t]
       \caption{Components of questions from XiaoMu users. \textit{N}, Tot \%, and Sub \% are short for number, the proportion of each type, and sub-type of questions.}
    \centering
    \small
    \begin{tabular}{c|c|c|c|c|c}
        \toprule
\textbf{Type} & \textbf{\textit{N}} & \textbf{Tot \%} &\textbf{ Subtype} & \textbf{\textit{N}} & \textbf{Sub \%}  \\
                \midrule 
         
  \multirow{2}{*}{ \shortstack{ Knowledge \\ Questions} } & \multirow{2}{*}{2410} & \multirow{2}{*}{60.22\%} & Simple & 2117 & 52.90\%  \\
  & & & Complex & 293 & 7.32\% \\
        \midrule 

  \multirow{2}{*}{ \shortstack{Chit-Chat} } & \multirow{2}{*}{702} & \multirow{2}{*}{17.54\%} &  General  & 571 & 14.27\%  \\  
  & & &  Emotional  & 131 & 3.27\% \\
         \midrule 

 \multirow{2}{*}{ \shortstack{Others} } & \multirow{2}{*}{890} & \multirow{2}{*}{22.24\%}& Platform FAQs & 449 & 11.22\%  \\
& & & Others & 441 & 11.02\%  \\
        \bottomrule

    \end{tabular}

    \label{tab:question_component}
\end{table}

\textit{Observation 1: Heterogeneous knowledge is vital to informative answers}. Simple questions are not \textit{simple}, but complex as they extend across different courses and domains, inside and outside of course contents. Therefore, only with heterogeneous knowledge sources, including course contents, search engines, and FAQs, as well as proper knowledge retrieving and curation methods, can VTAs generate concrete and informative answers.

\textit{Observation 2: A well-designed generation stage is essential.} Apart from simple questions that can be directly solved by retrieving methods, complex questions are also an important part of queries, which require special designs to generate instructive responses. Chit-chat should also be considered in the generative stage.

\subsection{Problem Formulation}

In this section, we first list the concept and formulation of heterogeneous resources and then define the task of VTA.


\textbf{Course Concept} is generally an academic term or
the non-academic category taught in the course video.

\textbf{Concept Graph} is a domain-specific knowledge graph with course concepts as head $k_h$ and tail $k_t$, the prerequisite between concepts as relation $r_p$, which is stored in a structured format $(k_h,r_p,k_t)$. Each concept has its explanation and domain.

\textbf{Search Engine} refers to online search API, which returns semi-structured top-k results for each query: $(q,[a_1,...,a_k])$.

\textbf{FAQ} is the frequently-asked questions with answers labeled by experts, which can be denoted as QA-pair format $(q,a)$.

\textbf{Virtual Teaching Assistant.} Consider the user's status consisting of the user's conversation history $x$ and learning status $s$. $x$ can be formally denoted as $x_t = \{q_1,y_1,..., q_{t-1}, y_{t-1},q_t\}$, where $q_i$ and $y_i$ is the $i$-th round query and response from the user and the VTA respectively and $q_t$ is the current query. $s$ is a group of information recording the user's learning progress, like the course $c$ the user attends. Distinct from conventional question answering or dialogue setting that aims at generating responses only based on $x$, the task of \textbf{
VTA} can be formulated as: given history $x_t$ and learning status $s$, the objective is to retrieve relevant knowledge $K_{x_t,s} = \textbf{f}(x_t,s)$ and generate knowledgeable response $y_t$:
\begin{equation}
   y_t  = \mathop{\arg\max}_{y}P(y|x_t, K_{x_t,s})= \mathop{\arg\max}_{y}P(y|x_t, \textbf{f}(x_t,s))
\end{equation}
where function \textbf{f}(·) denotes the model's ability to select appropriate knowledge from information sources. Because of the complexity of the user's queries, an ideal VTA should be proficient in understanding the user's intention (i.e. query for knowledge or chit-chat) and powerful in \textbf{f} to be accurate and informative.

\section{LittleMu System}

\begin{figure*}[htbp]
    \centering    \includegraphics[width=\linewidth]{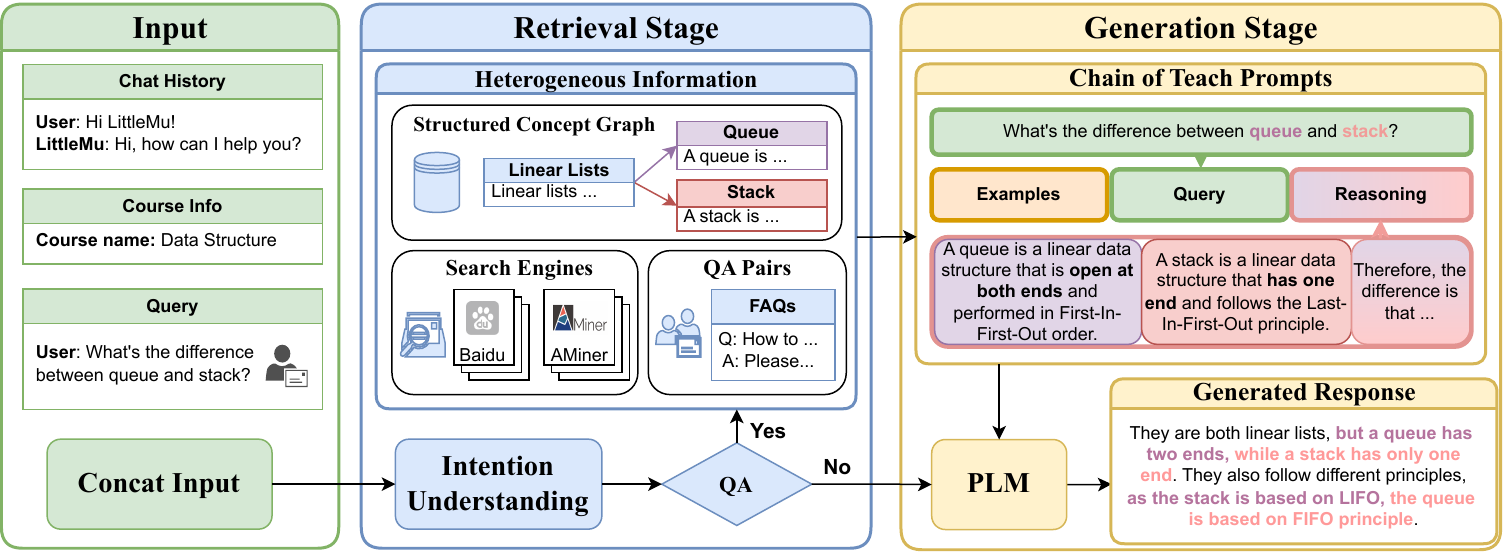}
    \caption{ 
    The two-stage framework of our \modelnames model. For chit-chat, \modelnames directly utilize the PLM to generate responses. For complex questions, \modelnames retrieves knowledge from heterogeneous information sources, then integrate the knowledge as reasoning to generate instructive responses. The Examples block stands for collected Chain of Teach examples.
    }
    \label{fig:pipeline}
\end{figure*}

\subsection{Overview}
As shown in Figure \ref{fig:pipeline}, our \modelnames system adapts a 2-stage structure: retrieval and generation. 

(1) \textbf{Retrieval stage}: This stage is responsible for retrieving knowledge snippets from heterogeneous information sources, which can answer those simple questions. It collects knowledge texts from heterogeneous sources to build the repository for candidate responses offline and selects candidate knowledge snippets according to the user's intention online. The retrieved snippet is either returned as the response to simple knowledge questions or passed to the generation stage for knowledge-injected prompting.

(2) \textbf{Generation stage}: This stage focuses on generating the novel and fluent response as a supplement for retrieved responses. Furthermore, we design two knowledge-guided prompting methods to generate knowledge-grounded chit-chat responses and answer complex questions with explainable reasoning processes.








Next, we will introduce both modules and their interaction.

\subsection{\retrievalname}
\label{sec:knowledge_intergaration}



\subsubsection{Intention Understanding}
To understand the user's intention, we employ an intention classification module to distinguish QA and chit-chat queries. which adopts an ALBERT~\cite{lan2019albert} model with a linear layer for prediction, i.e.,

\begin{align}
    d &= \mathtt{[CLS]}x_{t-1},c\mathtt{[SEP]}q_t\\
    \mathbf{d} &= \text{ALBERT}(d) \\
 \mathbf{h} &= \text{softmax}(\text{linear}(\mathbf{d})) 
\end{align}
where  $\mathbf{d}$ is the representation vector for the concatenated input $d$ and $\mathbf{h}$ is the predicted chit-chat intention score.  Here we set a threshold $\alpha$ to control the intention classification. If $ \mathbf{h} > \alpha$, then we will classify the query as chit-chat and enter the generation stage directly. Otherwise, we will continue the retrieval stage to collect and rank the relevant knowledge resources for the user's question.

\subsubsection{Knowledge Curation}


As students need fine-grained knowledge, we adapt the data collection framework of MOOCCubeX~\cite{yu2021mooccubex} to build a \textbf{concept graph} for the MOOC domain. We collected corpus from the MOOC video subtitles on XuetangX, which contains rich concept information about the course. As human annotation is labor-intensive, we employ a weakly supervised concept extraction and prerequisite discovery pipeline to obtain the entity and relation on the concept graph. 
As MOOC covers various categories of knowledge, we should also 
collect from open-domain \textbf{search engine} as a supplement. 
We choose Baidu as the search engine to retrieve up-to-date world knowledge. We also leverage AMiner~\cite{tang2016aminer} as the academic search engine to provide information about scholars, papers, and research trends as an extension to MOOC content. 
In addition to the structured concept graph and semi-structured search engine, we also construct a QA-formatted \textbf{FAQ}  database and maintain periodic updates. \modelnames provides the option of ``asking real TA" for users, these queries will be cached and then reviewed by experts. The human-written answers are recorded in the FAQ database as an information source. 

Since the knowledge resources have heterogeneous formats, we transform them into a unified QA-pair format: the concept on the structured concept graph is extracted as (concept, explanation) pair. 
Besides, the semi-structured snippets from the search engine are flattened into (headline, text) pairs. The FAQ information is already in QA-pair format. Finally, we index all these QA-pair format data by Elasticsearch~\cite{gormley2015elasticsearch} search engine to facilitate the calculation of BM25 scores in the following steps.

\subsubsection{Concept-aware Ranking}
To solve those simple questions seeking specific knowledge, 
which are mostly information of certain concepts 
, it is time-efficient to find answers by information retrieval. With the heterogeneous resources, \modelnames can ensure the recall of retrieved snippets. Furthermore, we propose a concept-aware metric $\mathbf{S}$ based on BM25 with heterogeneous weights, which ranks the candidate snippets $Z$ to improve the precision. 
Considering the user's learning status and intention, we will reward those candidate concepts $\mathbf{K}$ from the course that the user is learning when the user is seeking accurate concept knowledge, we will also give higher weights to those search engine snippets $\mathbf{E}$ when the user intends to ask an open question:

\begin{align}
    \mathbf{g}(z) &= \frac{| D(c_z)\cap D(c)|}{ | D(c_z)\cup D(c) |}\\
    \mathbf{S}(z,q) &=\left\{
\begin{matrix*}[l]
\mathbf{g}(z) \cdot \text{BM}25(z, \text{NER}(q)) & \quad \quad \text{\textit{if}}\quad z\in \mathbf{K}\\
\mathbf{h} \cdot \text{BM}25(z, \text{NER}(q)) & \quad \quad \text{\textit{if}}\quad z\in \mathbf{E}\\
\text{BM}25(z, \text{NER}(q)) & \quad \quad \text{\textit{otherwise}}
\end{matrix*}
\right.
\end{align}
where $c_z$ is the course of the candidate concept $z$ belongs to, $NER(q)$ are extracted key concepts from the query, and $\mathbf{g}$ is the Jaccard similarity between the retrieved concept's course domain $D(c_z)$ and the user's current learning course's domain $D(c)$. 



Finally, we rank heterogeneous sources above by their scores. If one of the top $K$ retrieved snippets has a high probability conditioned on the current user's query and course, then we directly return the answer to the user. Specifically, we employ the concept-aware ranking metric  $\mathbf{S}$  to model the retrieval probability  $ P_{\theta}\left(z\middle|x,c\right)$:
\begin{equation}
topK(Z)= \mathop{\arg\max}_{z\in Z}\left(P_{\theta}\left(z\middle|x,c\right)\right) \propto  \mathop{\arg\max}_{z\in Z}(\mathbf{S}(z, x, c))
\end{equation}
where 
\modelnames sets a threshold $\beta$ to measure whether these snippets match the user's query. If the ranking score $ \mathbf{S} > \beta$, then we can assume the answer is contained in the retrieved snippet.

\subsection{\generationnames}

\subsubsection{Chain of Teach for Complex Questions} Other than the simple knowledge questions, 
there are some questions that require reasoning over concepts rather than just utilizing the retrieved unstructured knowledge without an explicit reasoning process. 
As analyzed in section \ref{sec:question_analysis}, 
These questions generally require analyzing the relation and extension over concepts, while the retrieved knowledge only contains facts and basic concept explanations. To solve these complex questions, we exploit PLM's few-shot reasoning ability via a novel prompting method.

Following the idea of the Chain of Thought prompting~\cite{wei2022chain}, which inserts explicit reasoning process examples that guide the PLM to answer complex questions step by step, we propose a Chain of Teach algorithm to provide concept explanations, prerequisites, and domain information for the users' queries. As most MOOC learners' questions are related to the concepts mentioned in the course, providing answers with concepts' explanations may help learners better understand the course's knowledge. For example, when students ask ``\textit{What's the difference between stack and queue?}", teachers will first explain what stack and queue are, then analyze their difference. We ask a group of expert teachers to write a collection of explaining chain examples for the Chain of Teach structure. Besides, to grasp the knowledge structure of concepts, the prerequisite concepts and the belonging domains for the mentioned concepts are also useful. The process is described in Algorithm \ref{algorithm:cot}. Notice that if there are no concepts extracted from the query, our algorithm will degrade to the standard Chain of Thought Algorithm. 

\begin{algorithm}[!ht]
  
	\caption{Chain of Teach}
	\textbf{Input: }{
	The concept graph $G_c$ for current course $c$.}\\
	{\hspace{0.89cm} The collection of Chain of Teach examples, $S_{cot}$.}\\
	{\hspace{0.89cm} Pre-trained Language Model, PLM.}\\
	{\hspace{0.89cm} The user's query, $q$.}\\
	\KwOut{The answer $a$ and the reasoning prompt $r$ }
        $\{k_1,k_2,...,k_m\} \leftarrow \text{NER}(q)$ \Comment{Concept Extraction} \\
         $i \leftarrow 1, r \leftarrow \text{`` "}$ \Comment{Initialization} \\
		\While{$i \leq m$}{
		    $r \leftarrow r +  k_{i}.\text{definition} $ \Comment{Concept Explanation}\\
		    $ D_i  \leftarrow  \text{find\_domain} (G_c,k_i)$ \\
			$r \leftarrow r + \text{``} k_i \text{ belongs to domain } D_i \text{"}$\\
			$ \{k_{i1},k_{i2},...,k_{in}\}  \leftarrow  \text{find\_prerequisite\_concept} (G_c,k_i)$\\
			$r \leftarrow r + \text{``The prerequisite concepts of } k_i \text{ are:"}$\\
			$j \leftarrow 1$ \\
				\While{$j  \leq n$}{
						$r \leftarrow r +  k_{ij}.\text{definition} $ \Comment{Add prerequisites}\\
						$j \leftarrow j+1$\\
				}
			$i \leftarrow i+1$\\
		}
	 
	 $e  \leftarrow  \text{sample\_similar}(S_{cot}, q)$ \Comment{Add example prompt} \\
	 $f \leftarrow  e + q + r $  \Comment{Final prompt for PLM}  \\
	$a \leftarrow  \text{PLM}(f) $  \Comment{Generate answer by PLM}  \\
	\Return $a, r$
	
	  \label{algorithm:cot}
\end{algorithm}

\subsubsection{Generative Method for Chit-Chat} As observed in section \ref{sec:question_analysis}, there are 18\% queries aiming to just have a chat. To satisfy the social interaction need of students, we provide a chit-chat service. Following XDAI~\cite{yu2022xdai}, we utilize the background knowledge underlying the dialogue histories to build prompting templates that guide PLM to generate chit-chat responses.





\section{Experiment}

We conduct experiments via both human and automatic evaluation to analyze \modelname. 

\subsection{Experimental Settings}
\begin{table}[t]
   \caption{Statistics of the test data for each task. Courses, Participants, and Label are respectively the number of background MOOCs, annotators, and collected dialogue labels. }
    \centering
   \small
    \begin{tabular}{c | c c  c}
        \toprule
  \textbf{Task} & \textbf{Courses} &  \textbf{Participants} &  \textbf{Label}   \\
        \midrule 
 Question Answering &   108  & 10 &   1,767 \\
    General Dialog   &    20  & 20 &   124,512 \\
        \bottomrule
    \end{tabular}
 
    \label{tab:test_statistics}
\end{table}

\begin{table*}[htbp]
       \caption{Human evaluation results of different models under the educational dialogue settings with the 95\% confidence intervals.  }
    \centering
    \small
    \begin{tabular}{c|c|c c c c c c }
        \toprule
\multirow{2}{*}{ \textbf{Category}}    &    \multirow{2}{*}{ \textbf{Model}} & \multicolumn{6}{c}{ \textbf{General Dialogue Quality}} \\
        
    &     &  \textbf{Coherence} &  \textbf{Informativeness} &  \textbf{Hallucination} &  \textbf{Humanness} &  \textbf{Helpfulness} &  \textbf{Instructiveness}  \\
                \midrule 
         
  \multirow{3}{*}{ \shortstack{Pre-trained \\ Language \\ Model} } & CPM-2~\cite{zhang2021cpm}  & 0.17$\pm$0.01 & 0.22$\pm$0.01 & 0.15$\pm$0.01 & 0.58$\pm$0.02 & 0.13$\pm$0.01 & 0.09$\pm$0.01  \\
  & GLM~\cite{du-etal-2022-glm}  & 0.98$\pm$0.04 & 0.94$\pm$0.04 & 0.95$\pm$0.04 & 1.20$\pm$0.03 & 0.89$\pm$0.04 & 0.65$\pm$0.03 \\
  & GLM-130B~\cite{zeng2022glm}  & 1.50$\pm$0.03 & 1.45$\pm$0.03 & 1.57$\pm$0.03 & \textbf{1.55$\pm$0.03} & 1.42$\pm$0.03 & 1.02$\pm$0.03 \\
        \midrule 

  \multirow{3}{*}{ \shortstack{Open-domain\\ Dialogue \\Model} }  &  CDial-GPT~\cite{wang2020large}  & 0.47$\pm$0.03 & 0.44$\pm$0.03 & 0.44$\pm$0.03 & 0.95$\pm$0.04 & 0.35$\pm$0.03 & 0.22$\pm$0.02  \\  
  &  EVA2.0~\cite{gu2022eva2}  & 0.81$\pm$0.04 & 0.69$\pm$0.04 & 0.76$\pm$0.04 & 1.01$\pm$0.04 & 0.59$\pm$0.03 & 0.47$\pm$0.03 \\
    &   PLATO-XL~\cite{bao2021plato} & 0.92$\pm$0.04 & 0.87$\pm$0.03 & 1.06$\pm$0.03 & 1.47$\pm$0.03 & 0.67$\pm$0.03 & 0.55$\pm$0.03  \\

         \midrule 

 \multirow{3}{*}{ \shortstack{Virtual\\ Teaching   \\Assistant} } 
& XiaoMu~\cite{Song2021XiaoMuAA} & 1.27$\pm$0.04 & 1.25$\pm$0.04 & 1.26$\pm$0.04 & 0.66$\pm$0.03 & 1.16$\pm$0.04 & 0.72$\pm$0.03  \\

 & \modelnames(10B)  & 1.46$\pm$0.03 & 1.46$\pm$0.03 & 1.50$\pm$0.03 & 0.89$\pm$0.04 & 1.35$\pm$0.03 & 1.05$\pm$0.03  \\
 & \modelnames(130B)  & \textbf{1.60$\pm$0.05} & \textbf{1.59$\pm$0.05} & \textbf{1.59$\pm$0.05} & 1.07$\pm$0.07 & \textbf{1.55$\pm$0.05} & \textbf{1.45$\pm$0.06} \\
   
        \bottomrule
    \end{tabular}

    \label{tab:general_dialog_quality}
\end{table*}

\paragraph{\textbf{Data Collection}}

To facilitate the development and evaluation of future VTA systems, we collect and label a dataset for both QA and dialogue tasks. The dataset's statistics are shown in Table \ref{tab:test_statistics}. We first sample 4002 
 user queries and annotate their intentions. Then, for those queries with QA intention, we ask the annotators to write an answer as a reference. The models for QA tasks are tuning-free so we use all the data for test. Apart from QA, we also conduct a human evaluation on the general dialogue task, where the participants can ask the VTA anything under the setting that they are learning a certain MOOC on XuetangX. To simulate the real MOOC learner interaction, we recruit 20 people, mainly university students, to generate conversations and score the dialogue quality. 


\paragraph{\textbf{Evaluation Metrics}} We assess \modelname's performance via both automatic metrics and human evaluation.

\textit{Question Answering}: We use several automatic metrics to evaluate \modelname's QA ability on our labeled test set. Following previous works on open-domain and conversational QA~\cite{anantha-etal-2021-open,christmann2022convmix},  we use \textbf{ROUGE}~\cite{lin2004rouge} to evaluate the overlap between the generated answer and the reference answer.

\textit{General Dialogue}: We examine how \modelnames performs in dialogue tasks generally via human evaluation in six aspects, following knowledge-grounded dialogue works~\cite{yu2022xdai, huang-etal-2021-plato} and special designs in teaching assistant context: (1) \textbf{Coherence} is to measure response's consistency and relevance with context. (2) \textbf{Informativeness} is to measure whether the response is informative. (3) \textbf{Hallucination} is a fine-grained metric for informativeness evaluation, checking the correctness of the fact. (4) \textbf{Humanness} evaluates how much the chatbot seems like a human being. Considering \modelnames as a teaching assistant chatbot, we also employ (5) \textbf{Helpfulness} and (6) \textbf{Instructiveness} to evaluate whether the response answers students' query and provide instructive information. The scale is \{0,1,2\} in the 6 aspects, and higher score indicates a better performance.

\paragraph{\textbf{Baselines}}  For the QA and general dialogue evaluation, we reproduce three representative categories of baseline models or call their APIs for comparison: (1) Generative Pre-trained  Language Models ~\cite{zhang2021cpm,du-etal-2022-glm,zeng2022glm}, (2) Open-domain Dialogue Models ~\cite{wang2020large,gu2022eva2,bao2021plato} and (3) Retrieval-based QA Model~\cite{hsu2022xiao,qiu2022dureader_retrieval,chen-etal-2017-reading}. Considering that \modelnames is deployed on a commercial platform XuetangX, we mainly choose open-source models for comparison. Besides, we compare different versions of \modelnames. The first version is called \textbf{XiaoMu}~\cite{Song2021XiaoMuAA} and has been deployed on XuetangX with continuous development since May 2020. Furthermore, we utilize PLMs to build the 
prompt-based generation module and combine it with the retrieval module to form the \modelnames system in July 2022. For the ablation study, we tried two kinds of PLMs: GLM~\cite{du-etal-2022-glm} and GLM-130B~\cite{zeng2022glm}, which correspond to the \textbf{\modelnames(10B)} and \textbf{\modelnames(130B)} versions.

\subsection{Result Analysis}

\subsubsection{Experiment for General Dialogue Quality}
\label{sec:general_dialogue}

We compare \modelname's performance by human evaluation with other dialogue models. Each model generates 1000 responses in 20 different courses, 400 of whose queries are the same, and 600 of whose queries are generated by volunteers. To avoid the influence of preference, every conversation is labeled twice by different annotators, and the final score is determined by average. The results are summarized in Table \ref{tab:general_dialog_quality}, which demonstrate following phenomenons:

(1) \modelnames outperforms other models in almost every dimension, especially helpfulness and instructiveness, which demonstrates that \modelnames is more effective in solving students' diverse problems and assisting them to learn, because of its enriched knowledge and special design in MOOC scenarios.

(2) Compared with its old version XiaoMu, \modelname's dialogue quality improves significantly, benefiting from its extra generation stage for chit-chat and complex question. 

(3) \modelnames still needs to promote itself in talking like a human. As a VTA rather than a chatbot, it might affect its score that \modelnames prefers to generate responses with more information and knowledge rather than fluent but trivial sentences like \textit{I see}, \textit{I don't know}, while the latter seems more humanlike than the former. 

\begin{figure}[htbp]
\centering
\subfigure[Performance on 5 course categories.]{
\begin{minipage}{0.49\linewidth}
\centering
\includegraphics[width=\linewidth]{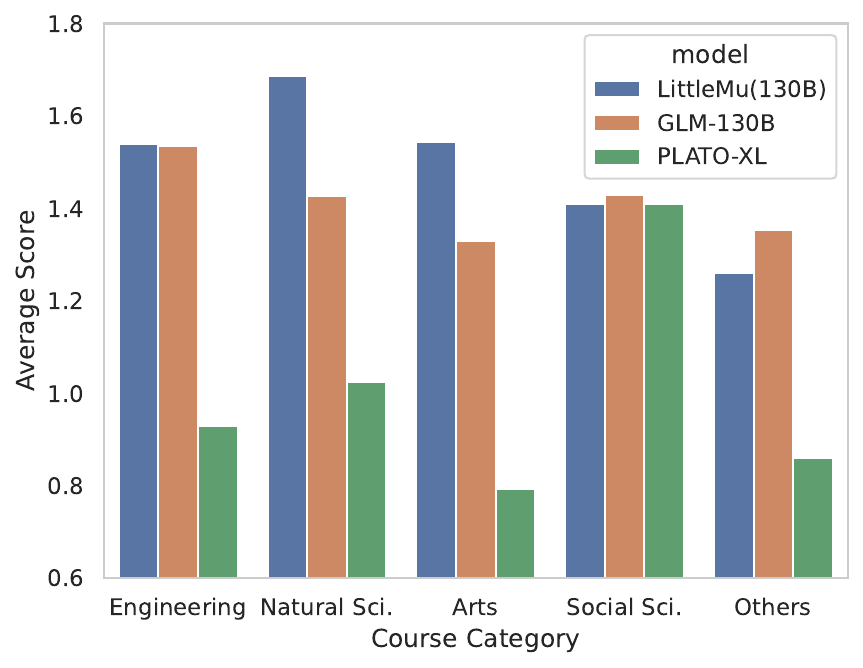}
\label{fig:course_score}
\end{minipage}%
}%
\subfigure[The average number of session rounds and satisfaction rate of user interaction with \modelnames over periods.]{
\begin{minipage}{0.49\linewidth}
\centering
\includegraphics[width=\linewidth]{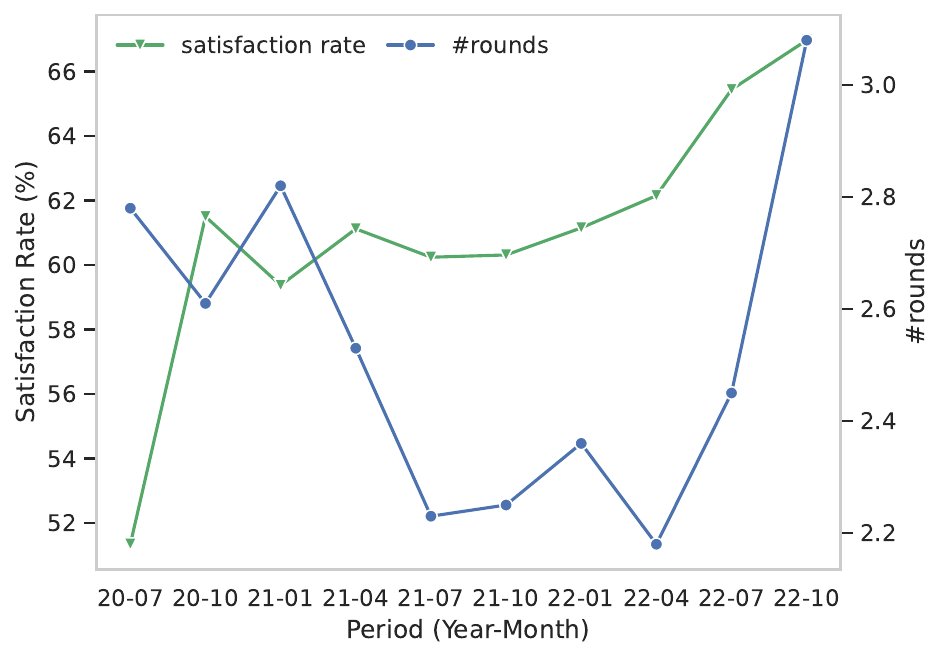}
\label{fig:session_rounds}
\end{minipage}%
}%

 
\centering
\caption{Domain adaption results and online deployment statistics, where Sci. is short for Science course. }
\end{figure}

We also divide examined courses into 5 categories: \textit{engineering}, \textit{natural science}, \textit{arts}, \textit{social science} and \textit{others}, to compare models' performance between different knowledge contexts. For each kind of baseline, the best model's score is in Figure  \ref{fig:course_score}. 

(1) Among all categories, \modelnames performs best in \textit{engineering}, \textit{natural science}, and \textit{arts} courses, and is significantly superior in the latter two. GLM-130B is better in \textit{social science} and \textit{others}.

(2) Due to the Chain of Teach method that promotes its reasoning capacity, \modelnames is able to outperform others in \textit{engineering}, \textit{natural science}, and \textit{arts}, while GLM-130B distinguishes itself in \textit{social science}, probably because it was trained with large-scale corpus.

(3) Without extra training on specific courses, \modelnames maintains a satisfying performance on different kinds of courses, demonstrating its strong transfer ability as a generalizable VTA.


\begin{table}[t]
   \small
       \caption{Automatic question answering evaluation results of different models 1767 real questions asked by users. R1/2/L are corresponding ROUGE-1/2/L F1 scores with the reference.
    }
    \centering
    \begin{tabular}{c|c|c c c}
        \toprule
\multirow{2}{*}{\textbf{Category}}    &    \multirow{2}{*}{\textbf{Model}} &  \multicolumn{3}{c}{\textbf{ROUGE F1}} \\
        
    &     &\textbf{R1} & \textbf{R2} & \textbf{RL}  \\
                \midrule 
         
  \multirow{3}{*}{ \shortstack{Pre-trained \\ Language \\ Model} } & CPM-2~\cite{zhang2021cpm} & 10.4 & 0.4 & 8.8 \\
  & GLM~\cite{du-etal-2022-glm}  & 12.6 & 1.5 & 10.1 \\
  & GLM-130B~\cite{zeng2022glm}  &  19.8 & 4.3 & 14.5 \\
        \hline 
 \multirow{3}{*}{ \shortstack{Retrieval-based\\ QA \\ Model} }  & Xiao-Shih~\cite{hsu2022xiao} &  17.5 & 7.4 & 13.8 \\
   & DPR + MRC~\cite{qiu2022dureader_retrieval}  &  7.2 & 1.8 & 6.3 \\
  &  BM25 + MRC~\cite{chen-etal-2017-reading} & 2.5 & 0.2 & 1.5 \\

         \hline 

 \multirow{3}{*}{ \shortstack{ Virtual\\ Teaching   \\Asistant} } 
& XiaoMu~\cite{Song2021XiaoMuAA} & 21.5 & 11.1 & 17.8 \\

 & \modelnames (10B) &  22.2 & 11.2 & 18.3 \\

  & \modelnames (130B)  & \textbf{22.4} & \textbf{11.3} & \textbf{18.5} \\
   
        \bottomrule
    \end{tabular}

    \label{tab:complex_qa_domains}
\end{table}

\subsubsection{Experiment for Question Answering}

We analyze the results of different models on our labeled QA dataset to estimate \modelname's performance on both simple and complex questions. \modelnames and XiaoMu have a great advantage over other models in \textit{simple questions} and \textit{all}, proving their utilization of heterogeneous knowledge sources to be effective, which equipped them with strong capability to answer the major part of students' queries.
From XiaoMu to \modelname, applying the Chain of Teach method enhances the model's reasoning ability on answering complex questions. 

\subsubsection{Online Evaluation} To prove the effectiveness of our optimization, we conducted an online satisfaction testing~\cite{huang2022DuIVA} to evaluate the performance. Specifically, experts from XuetangX have been labeling whether the system's answers can satisfy users every day since 2020. We also held periodic meetings with this expert to discuss the development direction of the system.  As shown in Figure \ref{fig:session_rounds}, the average number of dialogue rounds and satisfaction rate increased rapidly after the deployment of the generation module (July 2022), which indicates the growing user involvement of our system as the continuous development for dialogue skills. 

\section{Conclusion}

In this paper, we present a virtual MOOC teaching assistant, \modelname, for providing question answering and chit-chat for students on MOOC platforms. To reduce the dependency on labeled data, \modelnames retrieves from heterogeneous open-accessible information sources with concept-aware ranking metric and constructs Chain of Teach prompts to help language models generate specific responses. We conduct both human and automatic evaluations to compare \modelnames with other open-source models. We hope our few-shot workflow can provide an easy-to-transfer and data-efficient solution for developing virtual teaching assistants on MOOC platforms.

\section*{Acknowledgement}
This work is supported by the New Generation Artificial Intelligence of China (2020AAA0106501) , National Natural Science Foundation of China (No. 62277033) and a grant from the Institute for Guo Qiang, Tsinghua University (2019GQB0003).

\bibliographystyle{ACM-Reference-Format}
\bibliography{sample-base}

\end{document}